\documentclass[11pt]{article}

\usepackage[preprint]{acl}

\usepackage{times}
\usepackage{latexsym}
 
\usepackage[T1]{fontenc}

\usepackage[utf8]{inputenc}

\usepackage{microtype}

\usepackage{inconsolata}

\usepackage[normalem]{ulem}

\usepackage{graphicx}
\usepackage{booktabs}
\usepackage{graphicx}
\usepackage{comment}
\title{Classifying several dialectal Nawatl varieties}

\author{
  \textbf{Juan-José Guzm\'an-Landa\textsuperscript{1}},
  \textbf{Juan-Manuel Torres-Moreno\textsuperscript{1,2}},
  \textbf{Miguel Figueroa-Saavedra\textsuperscript{3}},\\
  \textbf{Carlos-Emiliano Gonz\'alez-Gallardo\textsuperscript{2,1}},
  \textbf{Graham Ranger\textsuperscript{4}},
  \textbf{Martha Lorena-Avenda\~no-Garrido\textsuperscript{3}}
\\
  \textsuperscript{1}LIA/Avignon Université  (France),
  \textsuperscript{2}LIFAT/Université de Tours  (France),\\
  \textsuperscript{3}Universidad Veracruzana (México),
  \textsuperscript{4}ICTT/Avignon Université (France)
\\
  \small{
    \textbf{Correspondence:} \href{mailto:juan-manuel.torres@@univ-avignon.fr}{juan-manuel.torres@univ-avignon.fr}
  }
}

\begin{document}
\maketitle
\begin{abstract}
Mexico is a country with a large number of indigenous languages, among which the most widely spoken is Nawatl, with more than two million people currently speaking it (mainly in North and Central America). Despite its rich cultural heritage, which dates back to the 15th century, Nawatl is a language with few computer resources. The problem is compounded when it comes to its dialectal varieties, with approximately 30 varieties recognised, not counting the different spellings in the written forms of the language. In this research work, we addressed the problem of classifying Nawatl varieties using Machine Learning and Neural Networks. 
\end{abstract}

\noindent {\bf Keywords:} {\sl Nawatl, Dialectal varieties, Automatic classification, SVM, CNN, LSTM}

\section{Introduction}
Languages with few digital resources constitute a large proportion of the languages spoken in the world \cite{these-pi,abdillahi:hal-01311495}.
and are distributed across large geographical regions. However, for a variety of reasons (political, economic, or social), they have been technologically neglected in Natural Language Processing (NLP). Consequently, these languages have low —or extremely low— computational resources and tools available for automatic processing.
%
By example, the Nawatl language have existed for thousands of years. It is one of Mexico's national indigenous languages with the second highest number of speakers, at approximately 1.65 million people \cite{inegi2020censo}, however Nawatl is actually a low-resource language. The language has considerable dialectal diversity categorised into four main geographical groups: Western, Central, Eastern, and Huastecan (Ethnologue, 2025\footnote{\url{https://www.ethnologue.com}}; \cite{Lastra1986areas, Canger1988NahuatlDA}).
The INALI\footnote{\url{https://www.inali.gob.mx/clin-inali/html/l_nahuatl.html}} (\textsl{Instituto Nacional de Lenguas Indígenas de México}) recognises 30 varieties of Nawatl distributed across different regions of Mexico. 

These groups are established according to phonological, lexical, and semantic features reflecting their variability, different processes of change, differentiation, and convergence, which are not unrelated to processes of contact between varieties and processes of specialization, intellectualism, and social prestige.

This linguistic diversity poses challenges for the use of corpora in educational, communicative, and digital contexts, as it involves significant variation in spelling and lexical selection. \cite{zimmerman,olko2016bridging,hansen2024nahuatl}.
This diversity, which reflects the historical and cultural evolution of Nahua-speaking communities over the last two centuries, is a condition that speakers themselves have viewed as a weakness. This idea is reinforced by the lack of a common standard register or variety. The result is a preference for Spanish as a common and apparently more uniform language. This linguistic attitude results in the loss of exchange value and use of the Nawatl language, leading to linguistic displacement. Thus, in cases of contact between speakers of different varieties, the tendency is to avoid interdialectal communication that requires mutual understanding and to opt for the exclusive use of Spanish. This, in turn, does not promote the possibility of new processes of standardisation and learning of language varieties, as in other linguistic communities. 
The process is all the more obvious in contexts of written communication. 
%
In the hope that new social and political conditions surrounding language policies will enable an increased use, acceptance and appreciation of mutual differences, the emergence of a multiplicity of cultural products in different varieties requires resources that will facilitate communication and enable these products to reach a larger target of consumers.

Thus, a NLP tool that could identify and classify different document types according to their linguistic and lexical characteristics would be a valuable resource in support of other tools, such as translators, emotion detectors, etc.

These tools empower and strengthen the community of users of this language by making information more accessible without having to give up the use of their mother tongue in order to access other variants. They can even be used as a learning resource to help users become familiar with other varieties of Nawatl, thereby enhancing the value of the cultural products generated from that language without having to mediate through Spanish. Such measures reinforce speakers' identification with the speech community, and highlight the usefulness and value of the language that defines it.
Our objectives are to adapt and combine various multi-class classifiers to identify different dialectal varieties of written Nawatl with high precision.

The rest of this paper is organized as follows: Section \ref{sec:nawatl} introduces Nawatl and its dialectal varieties. Section \ref{sec:piyalli} characterizes the corpus used in this study. Section \ref{sec:algos} presents the classification algorithms. Section \ref{sec:results} shows our results before concluding in Section \ref{sec:conclusion} and giving some directions for future work.

\section{Dialectal varieties of Nawatl}
\label{sec:nawatl}

Nawatl is an agglutinative and polysynthetic language, a feature which is formally manifest in the use of long words that concentrate multiple nuances in the expression of meanings. It can even give rise to veritable sentence-words, where a verbal word integrates both the subject and the predicate of a typical sentence in many Indo-European languages. This language belongs to the Yutonahuas language group. Originally located in the southwestern United States, Nawatl-speaking groups gradually moved southward in different waves, settling in territories that are now part of Mexico and Central America (Guatemala, El Salvador, Nicaragua). Thus, since the 5th century, Proto-Nahuan linguistic groups such as Pochuteco (now extinct) and Nawatl began to emerge, with the latter dividing into the western-central Nahua group and the eastern-northern Nahua group. Each of these groups came into contact with different environments and other linguistic and cultural groups. This historical development is reflected, in grammatical terms, in the great morphosyntactic and lexical-semantic variability that characterizes speech at regional and biocultural levels.
Given this reality, an automatic variety classification tool is becoming increasingly relevant. 

In Nawatl, there are varieties with a greater number of differences. 
This occurs, for example, between Huasteca Nawatl and Western Nawatl. Within these varieties, there are certain typological differences that do not in themselves represent a difficulty in understanding, as they are more morphological differences. It is observed that there is greater contrast between Western varieties (variety ``{\sc l}'') and Eastern (variety ``{\sc t}'') and these with the Central and Huasteca regions (varieties ``{\sc tl}''). In some cases, these variations can be difficult to understand and standardize among speakers and readers of different varieties, because the form of the word changes at the phonological, graphemic, and orthographic levels.

In this way, various forms can be found (for example {\it tlakatl, tlacatl, tagat, lakal}, etc. for ``male'' or ``man''; or {\it siual, sowatl, souatl, siuatl, siwat, cihuatl}, etc., for ``woman'', and also {\it tlamachtih, tlamachtihketl, temachti, temachtiani, tlamachtiani, tamaxtiani, lamaxtini}, for ``teacher'') which are semantically equivalent, varying only in form. However, when reading or hearing them differently, it is not always possible to understand that they are related.
Some variations are morphological in nature, relating to how words are constructed, whether nouns or verbs, in terms of agglutination. Thus, an example of the perfective and imperfective can be seen in Table \ref{tab:perfectivo}.

\setlength{\tabcolsep}{4.6pt} 
\begin{table*}
 \centering
 \begin{tabular}{|l|llllll|}
 \hline 
 & \bf Western & \bf Central & \bf Zongolica & \bf Huasteca & \bf Puebla & \bf El Salvador \\
 \bf Example & \bf MCO & \bf MCA & \bf CV & \bf HH HP HV & \bf & \bf NAW \\ \hline
 \bf I saw it &\it nikitak  &\it nikittak &\it nikittak &\it nikiitaki &\it nikitak &\it nikitak\\
 \bf I looked at him &\it nikchiak &\it onikchix &\it onikchixki &\it nikchixki &\it nikchi &\it nikchixki\\
 \bf I've got it &\it nikpia &\it nikpia & nikpiya &\it nihpiya &\it nikpia &\it nikpia\\
 \bf I had it &\it nikpiak &\it onikpix & onikpixki &\it nihpixki &\it nikpiak &\it nikpixki\\    
 \bf I saw it &\it nikitataya &\it nikittaya &\it onikittaya &\it nikiitayaya &\it nikitaya &\it nikitatuya \\
 \hline
 \end{tabular}
 \caption{\label{tab:perfectivo}
   Examples of perfective sentences.
 }
\end{table*}

Another case is, for example, the expression of the diminutive. Thus ``puppy'' can be found as {\it pilchichitsij, chichitzin, chichiton}, etc. Another example is ``grandparents'', that in some varieties it is said {\it tokolwan, tokokolwan o toweyitahwan}.
Thus, differences are not only found at the lexical or morphological level. They are also found at the semantic level. For example: “truth”, “for real”, “certainly”, can be expressed {\it nelli}  (Central) or {\it nelia} (Huasteca) o {\it nel} (Eastern) but also as {\it niman} (Western), when this word in other varieties is understood as “after”, “then” o “therefore”. Another example is the expression of ``my husband'', which can be said {\it nonamik} (``my partner'') or {\it noquich} (``my man''), but in other varieties {\it noueuej} (``my old man'').

A well-established example is the use of certain verbs. Thus, the verb “to be” can be expressed as {\it kah} or {\it yetok} in the Western and Central varieties, as {\it unkah} or {\it onkah} in the Nawatl of Guerrero, whereas in the Eastern and South {\it ono} is used, and in the Huasteca region, two forms are used: {\it eltok} for inanimate subjects and {\it itstok} for animate subjects. 
In this way, the phrase ``I am'' can be rendered as {\it nikah, nunkah, nono, niitstok}.
At the phraseological level, the varieties show real diversity and complexity. So, the sentence: ``There was a married man who had a wife" can be expressed as in Table \ref{tab:esposa}.

\setlength{\tabcolsep}{4.5pt} 
\begin{table*}[h!]
 \centering
 \begin{tabular}{|l|l|}
 \hline 
 \bf Regions & \bf ``There was a married man who had a wife'' \\ \hline
 \bf Western  & \it Niman kataya se lakal munamiktijtuk kipiataya isiua\\
 \bf Central   & \it Melak yokatka sentetl tlakatl monamiktitok kipiaya un isowah / \\
 & \it Nelli katki se tlakatl monamiktihtok okipiyaya in isiwaw\\
 \bf Huasteca &\it Nelia itstoya se tlakatl  kipiayaya ni isiwaj\\
 \bf Eastern &\it Nel onoya se tagat kipiaya monamiktitok ipalmiya /\\    
 &\it Nel nemik se takat munamiktijtuk kipiatuya ne isiwaw\\    
 \hline
 \end{tabular}
 \caption{\label{tab:esposa}
  Phraseological examples.
 }
\end{table*}

Thus, whether they display variations in form, construction, or meaning, it is extremely important to identify which varieties are involved. Hence the need for an efficient classifier of dialectal varieties. In this way, each variety will have a representation that can distinguish it from the others. We believe that this can be achieved using classical models of $n$-grams of characters. \cite{cavnar-1994-texcat} and also with dense representation models of word embeddings such as 
FastText \cite{bojanowski-etal-2017-enriching} 
combined with efficient Neural classifiers.

\section{Nawatl corpora}
\label{sec:piyalli}

Currently, there are few Nawatl corpora that are truly exploitable for NLP tasks. Among them is the corpus \textit{Axolotl} \cite{gutierrez2016axolotl}\footnote{\url{https://axolotl-corpus.mx}}, a parallel Nawatl-Spanish corpus with two Nawatl varieties: ``classic'' and ``modern'', but these two varieties are clearly unsuitable for our purpose. The $\pi$-yalli Nawatl corpus has been used in simple tasks of semantic textual similarity at the words and sentences level \cite{MICAI-piyalli-unigraph,piyalliTALN}. Another important achievement is the corpus of classical Nawatl, developed at UNAM.\footnote{Universidad Nacional Autónoma de México} and integrated into the ``Gran Diccionario N\'ahuatl''  (GDN)\footnote{\url{https://gdn.iib.unam.mx/}}.
Finally, there is the ``Universal Dependencies Treebank for Highland Puebla Nawatl''  corpus \cite{Universal_Depend_pugh-tyers}, which focuses only on the Sierra de Puebla Nawatl, with an extremely limited number of sentences.
It is clear that factors such as the oral nature of Nawatl, the lack of standardisation in spelling, and the existence of around thirty varieties mean that the number of available resources is not particularly high.

Among the different corpora, the $\pi$-yalli corpus has interesting features in the present context\footnote{See \url{https://demo-lia.univ-avignon.fr/pi-yalli}}. 
This is a heterogeneous --but small-- Nawatl corpus, that is suitable for training of non contextual embeddings.
However, this corpus has never been fully characterized, much less has any attempt been made to classify its different dialectal varieties. Then, we have decided to use the  $\pi$-yalli corpus in our experiments. 

Hereafter, the following acronyms will be used as labels for the 25 dialectal varieties available in the $\pi$-yalli corpus\footnote{In our study, we used the classification from the Atlas of Languages, Indígenas Nacionales de México (INALI) \url{https://atlas.inali.gob.mx/agrupaciones/info/0211}}:

\begin{description}
\item ~~~~{\bf CV}: Nawatl Central de Veracruz, {\bf H}: Huasteca\footnote{{\bf H} class is a heterogeneous mixture of varieties from the Huasteca region.}, {\bf HH}: Huasteca Hidalguense, {\bf HV}: Huasteca Veracruzana, {\bf HP}: Huasteca Potosina, {\bf CEA}: Mexicano del Centro Alto, {\bf CEO}: Mexicano Central de Occidente, {\bf CEN}: Mexicano del Centro,  {\bf CEB}: Mexicano Central Bajo, {\bf ORI}: Mexicano del Oriente, {\bf GUE}: Mexicano de Guerrero, {\bf NOC}: Noroeste Central, {\bf IST:} Nawatl del Istmo, {\bf NAW}: Nawat, {\bf OAX}: Náhuatl de Oaxaca,  {\bf CEP}: Centro de Puebla, {\bf ANP}: Alto del Norte de Puebla, {\bf ORP}: Mexicano del Oriente de Puebla, {\bf SNP}: Sierra Negra de Puebla, {\bf SNNP}: Sierra Negra Norte Puebla, {\bf SNSP}: Sierra Negra Sur de Puebla, {\bf SNEP}: Sierra Noreste de Puebla, {\bf SOP}: Sierra Oeste de Puebla, {\bf TEM}: Mexicano de Temixco,  {\bf TLA}: Mexicano de Tlaxcala.
\end{description}

\subsection{Corpus characterization}

The $\pi$-yalli corpus actually contains several millions of tokens\footnote{In the computer sense of the term: sequences of characters separated by spaces}. Often, each token represents a ``word-sentence'' in Nawatl.
%
The basic statistics of this corpus are as follows: 4~746 documents; $\approx$6~629~000 tokens; $\approx$428~000 sentences and 25 dialectal varieties.\footnote{$\pi$-yalli corpus is also divided into 16 thematic varieties, but this classification is beyond the scope of this paper.}
Documents containing numerous mixtures of dialectal varieties have been excluded from this study (e.g., Wikipedia documents), or those where the dialectal class could not be established with certainty.

\begin{figure}[h!]
  \includegraphics[width=1.05\columnwidth]{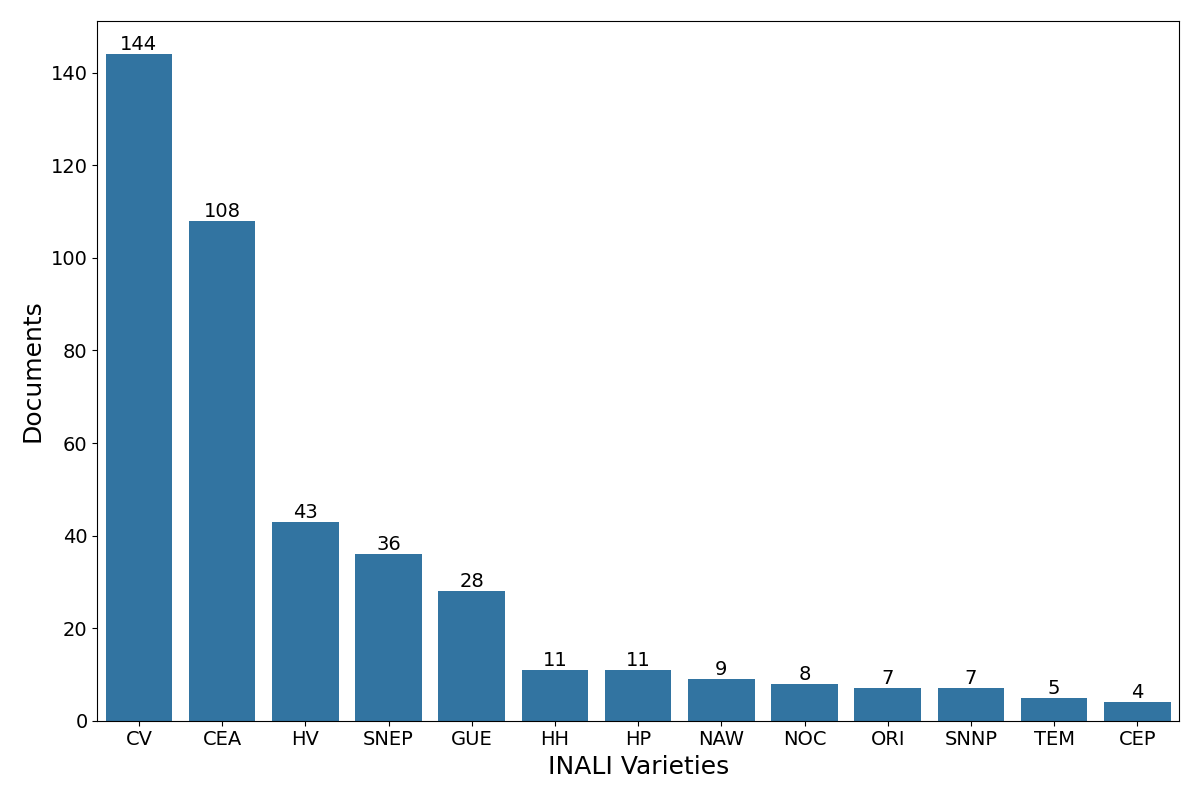}
  \caption{Number of documents concerning the 25 Nawatl varieties from $\pi$-yalli corpus. The varieties H, IST, OAX, CEO SOP, ORP, TLA, SNSP, CEB, ANP, SNP were excluded from the graph because they have fewer than three documents.}
  \label{fig:doc-var}
\end{figure}

Figure~\ref{fig:doc-var} shows the number of documents available for each dialectal variety.
As can be seen, the varieties ``Nawatl Central de Veracruz'' and ``Huasteca del Centro Alto'' dominate the others in terms of number of documents. 
At the other end of the spectrum, varieties such as {\bf SNP} ``Sierra Negra de Puebla'', {\bf ANP} ``Alto del Norte de Puebla'' or {\bf CEB} ``Mexicano Central Bajo'' are anecdotal in terms of number of documents.
 
\begin{figure}[h!]
\includegraphics[width=1.05\columnwidth]{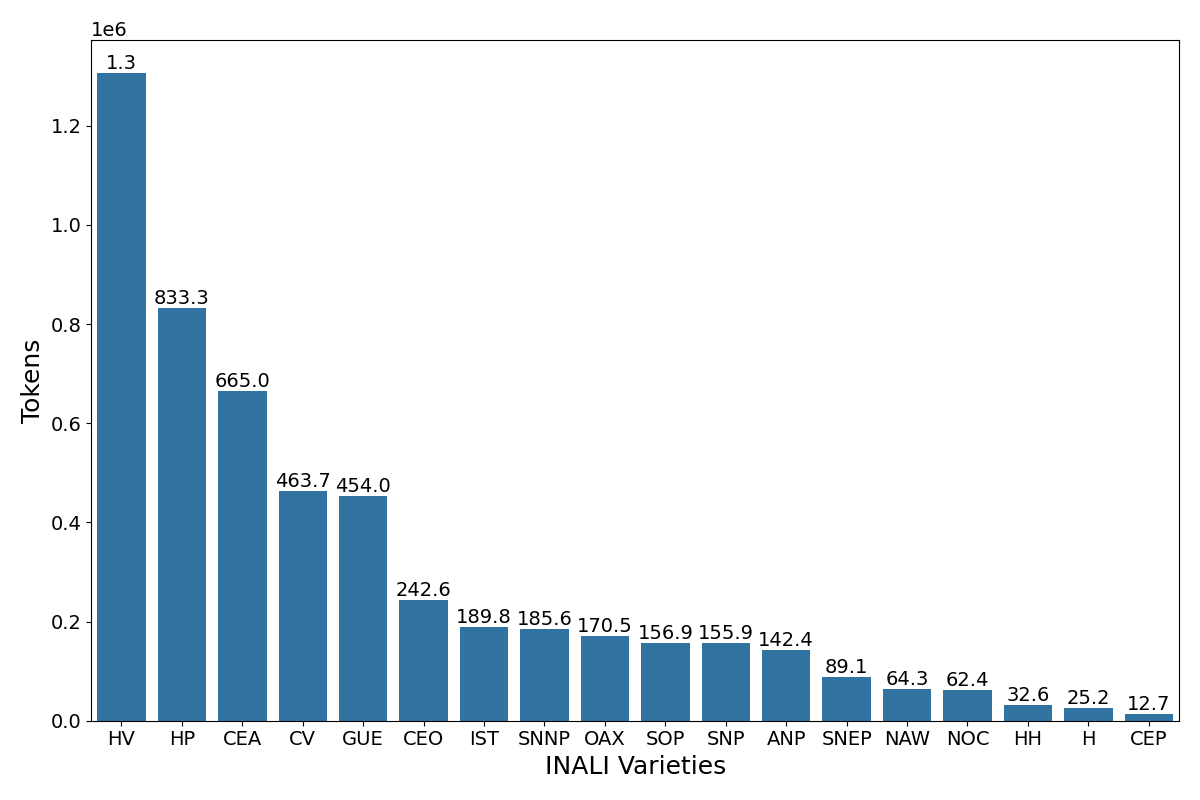}
  \caption{Tokens' number concerning the $k$=18 most representative varieties Nawatl from $\pi$-yalli corpus.}
  \label{fig:tok-var}
\end{figure}

Figure \ref{fig:tok-var} shows the number of tokens on the $k$=18 dialectal varieties from the $\pi$-yalli corpus.
Now the varieties ``Huasteca Veracruzana'', ``Huasteca Potosina'' and ``Mexicano del Centro Alto'' dominate the top three places. These first 18 varieties represent, in terms of the number of tokens, 99.7\% of all varieties in the corpus. The varieties ``Huasteca Hidalguense'', ``Huasteca'' or ``Centro de Puebla'' are barely represented.

\begin{figure}[t]
  \includegraphics[width=1\linewidth]{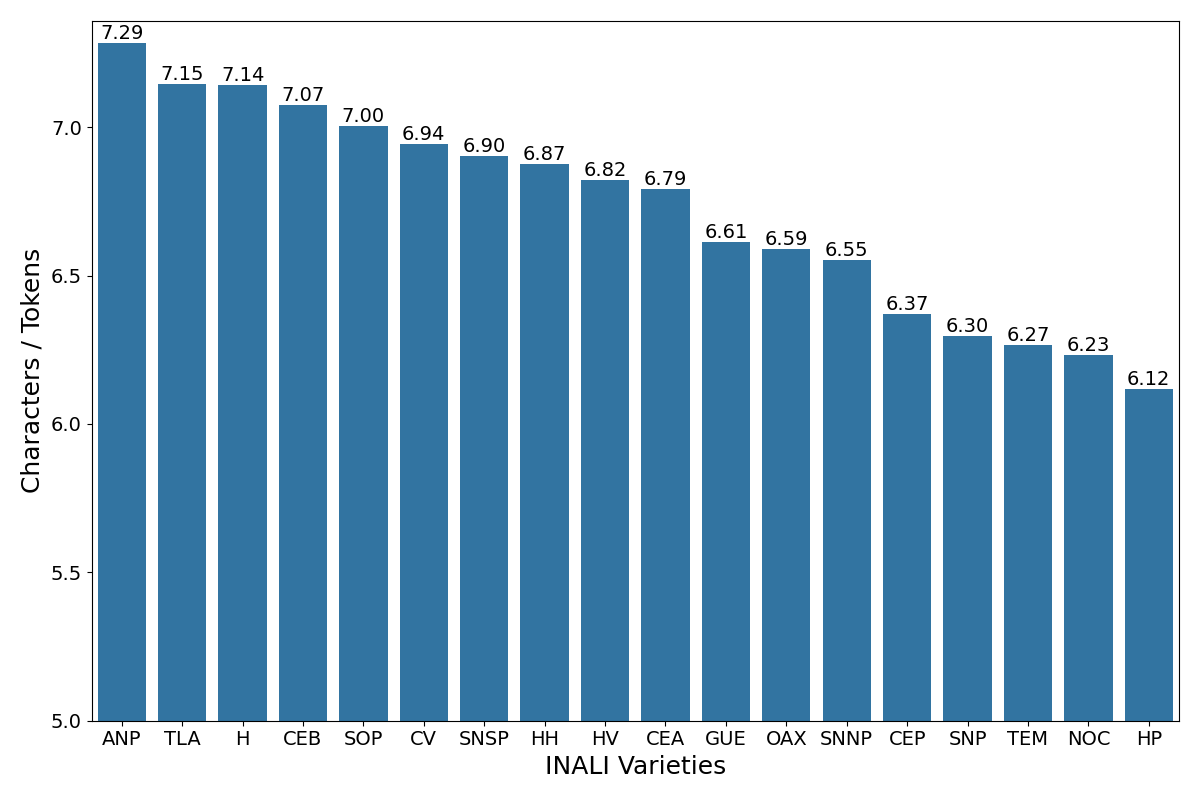}
  \caption {Average ratio between the number of characters and tokens in the Nawatl varieties of the $\pi$-yalli corpus. A token is an element divided between two spaces, but in Nawatl, the token can be a sentence thanks to agglutination. 
  \label{fig:car-tok}}
\end{figure}

However, the documents in this corpus vary greatly in size. In an analysis weighted by number of tokens and characters, the distribution of varieties shows a more balanced behaviour. For the Figure \ref{fig:car-tok} can be observed the average ratio between the number of characters and tokens. The variety {\bf HP} ``Huasteca Potosina'', is under-represented.
%

\begin{figure}[h!]
 \includegraphics[width=1.05\columnwidth]{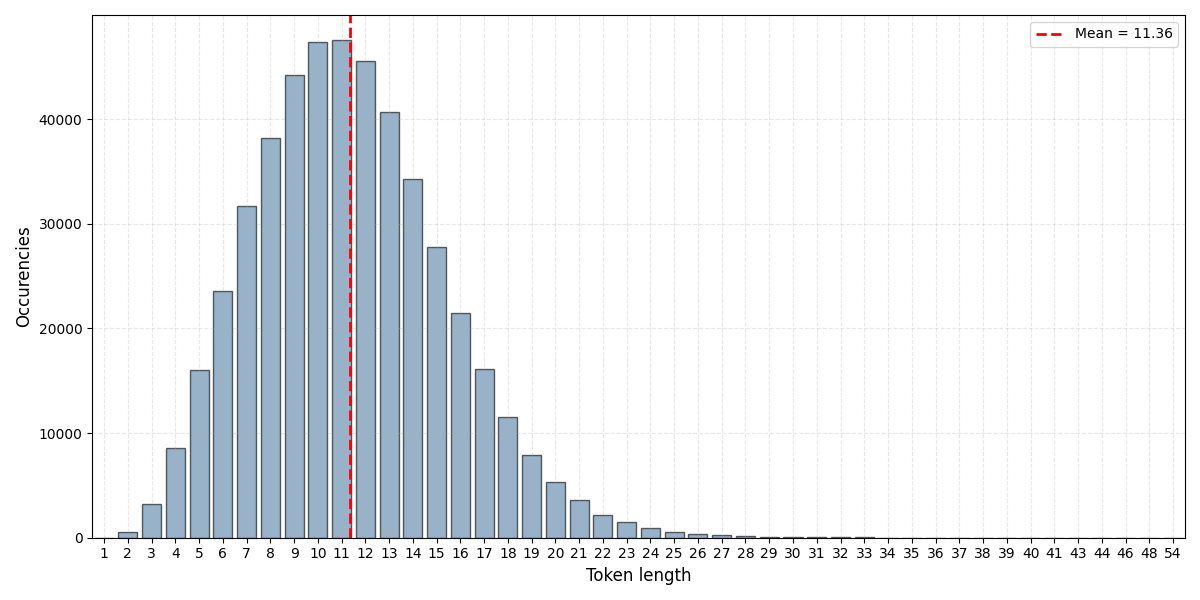}
 \caption{Token length in the $\pi$-yalli corpus.} 
 \label{fig:chars-tok}
\end{figure}

Finally, Figure \ref{fig:chars-tok} shows the number of characters per token (length of the ``word-sentences'') for the dialectal varieties studied. The average number of characters per token of 11.4 indicates a higher average word length than in Spanish (5.5), French (5.6), English (4.8), German (6.8), or Finnish (7.9)\footnote{\url{https://www.inter-contact.de/en/blog/text-length-languages}}. Nawatl even has a significant number of words with an average length of more than 15 characters. The article \cite{DBLP:journals/corr/abs-1208-6109} explores in greater depth the topic of word length using English and Russian over several centuries.


These statistical analyses show that the $\pi$-yalli corpus has a significant number of dialect classes, but that these are disproportionately distributed. 
The identification of its dialectal varieties can be seen as a potentially difficult classification problem. But is this really true? 
\subsection{Pre-processing}

All the documents in the $\pi$-yalli corpus are labelled according to their dialectal variety.
Based on Figure \ref{fig:tok-var}, we have decided to process only the first ones $k$=18 dialectal varieties with more than 10~000 tokens.
The texts were segmented into --reasonably-- identifiable sentences. Subsequently, sentences of length $|l|<5$ tokens, which produced $\approx$~325~000 sentences. The statistics for varieties and corresponding number of sentences are shown in Table~\ref{tab:nombre_phrases}.

\begin{table}[h!]
\centering
\begin{tabular}{|lll|}
\hline
\multicolumn{3}{|c|}{\textbf{Varieties: sentences}} \\
\hline 
{\bf HV~~~} : 92 283 & {\bf HP~~}: 57 310 & {\bf GUE}: 30 444 \\
{\bf CEA~~}: 22 831 & {\bf CV~~}: 19 725 & {\bf CEO}: 17 332 \\
{\bf SNNP}: 12 831 &{\bf ANP}: 12 106 & {\bf OAX}: 12 016 \\
{\bf SNP~~}: 11 273 & {\bf SOP}: 10,434 & {\bf IST~~}: 10 318 \\
{\bf NAW~}: 5 051 &  {\bf SNEP}: 4 518 & {\bf NOC}: 3 182 \\
{\bf H~~~~~~}  : 1 311 &  {\bf HH~~~~}: 1 113 & {\bf CEP}: 1 090 \\
\hline
\end{tabular}
\caption{Sentences/class. 18 class, $\approx$325K sentences.}
\label{tab:nombre_phrases}
\end{table}



\color{black}

\section{Classification algorithms}
\label{sec:algos}

The classification of dialectal varieties can be viewed as a multi-class classification NLP problem. These types of problems —when the number of classes is large— have a reputation for being difficult \cite{Manning:99,multiclass}. 
For example, \cite{Written-nahuatl-2021} have studied a problem with 10 varieties. Learning is performed by an LSTM network using 100-dimensional character embeddings on $\approx$7K sentences aligned by variety\footnote{The sentences are short verses of religious texts from \url{https://scriptureearth.org}}. The evaluation achieves 99\% accuracy but is limited to classifying a {\it single variety}, which we consider clearly insufficient. 

In our case, this is a realistic problem with $k$=18 varieties between which to discriminate.
Unfortunately, the limited size of the $\pi$-yalli corpus does not allow for the training of BERT-type LLM models, commonly used in multi-class text classification tasks. \cite{multiclass}. For this reason, the work is forced to use non-contextual embeddings and classical character representations.

In the experimental protocol, we used five Machine Learning algorithms to classify the Nawatl dialectal varieties:
\begin{itemize}
  \item Baseline using distances between character $n$-grams \cite{cavnar-1994-texcat};
  \item SVM large marge classifier \cite{SVM1995};
  \item LSTM Neural Network \cite{LSTM};
  \item Convolutional Neural Network CNN \cite{kim-2014-convolutional}:
  \item Siamese Neural Network C-LSTM \cite{CLSTM-Zhou}
\end{itemize}

\subsection{Baseline: $n$-grams of characters}

The baseline is a classic but powerful algorithm that uses character $n$-grams to identify languages. We use the TextCat algorithm. \cite{textcat} to calculate the proximity between varieties.
TextCat is efficient and has low computational costs because it uses few resources and computations. The algorithm establishes a notion of distance between language models of $n$-grams of characters with 2$\le n \le$5. In the test, the algorithm chooses the shortest distance between the $k$ tables of learned $n$-grams, calculating membership probabilities (computed on 2 000 $n$-grams). 

\subsection{Word embeddings}

In NLP, it is very important to have an adequate representations of text documents. Bag-of-Words and TF-IDF \cite{Manning:99} obtain intuitive and simple representations, but they lack context and suffer from high dimensionality problems in their embeddings.
On the other hand, more advanced techniques such as
FastText \cite{francis-landau-etal-2016-capturing} produce vector representations 
that capture important contextual information in reduced dimensions (CBOW or Skip-gram architectures). These embeddings have proven to be effective in textual similarity tasks in Nawatl \cite{MICAI-piyalli-unigraph}, for this reason, it was decided to use them in this work.

\subsection{Support Vector Machines}

Support Vector Machines (SVM) are highly efficient multi-class classification algorithms for high-dimensional vectors \cite{SVM1995}. The main idea is to find the optimal margin separating hyperplane. Soft margin hyperplanes are used in multi-class classification. SVM performs very well in multi-class classification tasks (with 23 and 90). \cite{textCategorization_SVM}. In both cases, the best result is obtained using Radial Basis Function (RBF).\footnote{Implementation used: \url{https://scikit-learn.org/stable/modules/svm.html}, with the parameters: kernel=rbf; regularisation=1.0; gamma=scale; MAX Iteration=unlimited.}

\subsection{Convolutional Neural Networks}

Convolutional Neural Networks (CNN) \cite{10.5555/2969033.2969120, 8308186} these are algorithms primarily focused on image processing.
In effect, convolution seeks to capture and regroup existing information in image segments.
However, textual information can also be processed in a similar manner. \cite{johnson-zhang-2017-deep, jacovi-etal-2018-understanding}. From textual segments of $m$ words, information can be extracted by applying convolution. \cite{kim-2014-convolutional} proposes that a filter consider a number of $m$ words within a sentence and the total number of features representing each word. The result of the convolution allows the information in a text to be condensed before passing to the layers of the fully connected network.\footnote{Pytorch was used. \cite{10.5555/3454287.3455008}, \url{https://pytorch.org} with the following parameters: filter sizes=3, 4, 5; dropout=0.5; \#~of filters=100; epochs=25 (using early stopping); learning rate=0.001. 
}

\subsection{Long Short-Term Memory} 

Long Short-Term Memory (LSTM) networks \cite{LSTM} are a type of recurrent neural network architecture with dedicated memories. 
LSTM works through three main components called gates: the forget gate, the input gate, and the output gate.
In this way, the model learns to retain important information despite having processed a large number of textual patterns \cite{9421225,JAMSHIDI2024102306}. 

\subsection{Siamese Networks} 

Siamese Neural Networks (C-LSTM) \cite{CLSTM-Zhou} are a type of architecture combining LSTM recurrent networks and CNN convolutional networks. In general, a CNN network is used before moving on to the LSTM architecture, with different numbers of filters and kernel sizes. In this work, 150 filters with kernels of dimensions 2, 3, and 4 were used. Siamese networks have been successfully tested in textual similarity tasks, \cite{tc-bi-lstm, siamese-TALN, 9407986, tsc-dpcnn}.

\section{Results}
\label{sec:results}
 
In this section, we show the results (Precision, Recall, F1-Score) of the multi-class classification algorithms for the 18 main Nawatl dialect varieties.
Cross-validation was performed (80\% training, 20\% testing, in 5 runs) to avoid undesirable bias effects. The Table \ref{tab:merged_all} summarises all the results obtained. 

\begin{table*}
\centering
\scriptsize
\setlength{\tabcolsep}{2.3pt}
\renewcommand{\arraystretch}{1.08}

\resizebox{\textwidth}{!}{%
\begin{tabular}{|l | ccc | ccc | ccc | ccc | ccc |}
\toprule
& \multicolumn{3}{c|}{\it \textbf{TextCat}} 
& \multicolumn{3}{c|}{\bf SVM} 
& \multicolumn{3}{c|}{\bf CNN} 
& \multicolumn{3}{c|}{\bf LSTM} 
& \multicolumn{3}{c|}{\bf C-LSTM} \\
\cmidrule(lr){2-4}\cmidrule(lr){5-7}\cmidrule(lr){8-10}
\cmidrule(lr){11-13}\cmidrule(lr){14-16}
\textbf{Variety}
& \textbf{P} & \textbf{R} & \textbf{F}
& \textbf{P} & \textbf{R} & \textbf{F}
& \textbf{P} & \textbf{R} & \textbf{F}
& \textbf{P} & \textbf{R} & \textbf{F}
& \textbf{P} & \textbf{R} & \textbf{F} \\
\midrule
HV   & 0.967 & \underline{0.499} & \underline{0.655}   & \bf0.976 & 0.735 & 0.838   & 0.964 & 0.973 & 0.969      & 0.971 & \bf0.977 & \bf0.974    & \underline{0.956} & 0.972 & 0.964     \\
HP   & 0.757 & \underline{0.935} & \underline{0.813}   & \underline{0.706} & \bf0.989 & 0.823   & 0.970 & 0.961 & 0.965      & \bf0.972 & 0.967 & \bf0.969    & 0.967 & 0.953 & 0.960     \\
GUE  & \underline{0.956} & \underline{0.823} & \underline{0.856}   & 0.991 & \bf0.995 & 0.993   & 0.986 & 0.992 & 0.989      & \bf0.994 & 0.994 & \bf0.994    & 0.990 & 0.991 & 0.991     \\
CEA  & \underline{0.821} & \underline{0.941} & \underline{0.852}   & 0.983 & 0.990 & 0.986   & 0.981 & 0.983 & 0.982      & \bf0.984 & \bf0.991 & \bf0.987    & 0.977 & 0.987 & 0.982     \\
CV   & \underline{0.785} & \underline{0.860} & \underline{0.806}   & 0.972 & 0.957 & 0.965   & 0.965 & 0.950 & 0.957      & \bf0.978 & \bf0.963 & \bf0.970    & 0.977 & 0.941 & 0.958     \\
CEO  & \underline{0.987} & \underline{0.994} & \underline{0.990}   &\bf 1.000 & 0.998 & \bf0.999 & 0.999 & 0.998 & 0.998     & 0.999 & \bf0.999 & \bf0.999    & 0.998 & 0.998 & 0.998     \\
SNNP & \underline{0.889} & \underline{0.804} & \underline{0.810}   & 0.977 & 0.981 & 0.979   & 0.964 & 0.974 & 0.969      & \bf0.981 & \bf0.983 & \bf0.982    & 0.968 & 0.971 & 0.969     \\
ANP  & \underline{0.903} & \underline{0.981} & \underline{0.939}   & \bf0.998 & 0.996 & \bf0.997   & 0.993 & 0.994 & 0.994      & 0.997 & \bf0.998 & \bf0.997    & 0.990 & 0.995 & 0.992     \\
OAX  & \underline{0.956} & \underline{0.954} & \underline{0.955}   & 0.988 & 0.991 & 0.989   & 0.983 & 0.986 & 0.984      & \bf0.990 & \bf0.993 & \bf0.991    & 0.983 & 0.981 & 0.982     \\
SNP  & \underline{0.969} & \underline{0.968} & \underline{0.968}   & \bf0.990 & 0.985 & 0.987   & 0.984 & 0.978 & 0.981      & 0.986 & \bf0.989 & \bf0.988    & 0.970 & 0.984 & 0.977     \\
SOP  & \underline{0.960} & \underline{0.944} & \underline{0.949}   & 0.986 & \bf0.997 & 0.991   & 0.985 & 0.992 & 0.988      & \bf0.991 & 0.995 & \bf0.993    & 0.980 & 0.994 & 0.987     \\
IST  & \underline{0.966} & \underline{0.992} & \underline{0.979}   & \bf0.999 & 0.998 & \bf0.999   & 0.997 & 0.996 & 0.996      & 0.998 & \bf0.999 & 0.998    & 0.997 & 0.997 & 0.997     \\
NAW  & \underline{0.991} & \underline{0.994} & \underline{0.992}   &\bf 0.999 & 0.997 & 0.998 & 0.997 & 0.998 & 0.997     & \bf0.999 & 0.998 & \bf0.999    & 0.995 & \bf0.999 & 0.997     \\
SNEP & \underline{0.600} & \underline{0.708} & \underline{0.592}   & \bf0.842 & 0.815 & 0.828   & 0.788 & 0.781 & 0.784      & 0.835 & \bf0.827 & \bf0.831    & 0.820 & 0.788 & 0.803     \\
NOC  & \underline{0.541} & \underline{0.713} & \underline{0.577}   & 0.851 & \bf0.950 & \bf0.897   & 0.827 & 0.864 & 0.844      & \bf0.858 & 0.882 & 0.870    & 0.836 & 0.916 & 0.873     \\
H    & \underline{0.418} & \bf0.960 & 0.517   & \bf0.847 & 0.588 & 0.694   & 0.704 & 0.578 & 0.625      & 0.828 & 0.684 & \bf0.744    & 0.723 & \underline{0.341} & \underline{0.429}     \\
HH   & \underline{0.519} & \underline{0.542} & \underline{0.414}   & \bf0.985 & 0.849 & 0.912   & 0.952 & 0.819 & 0.881      & 0.957 & \bf0.891 & \bf0.923    & 0.947 & 0.807 & 0.871     \\
CEP  & \underline{0.377} & \underline{0.792} & \underline{0.473}   & \bf0.967 & 0.859 & 0.910   & 0.946 & 0.835 & 0.885      & 0.949 & \bf0.911 & \bf0.929    & 0.909 & 0.803 & 0.851     \\
\midrule
\textbf{Mean}
& 0.799 & 0.857 & 0.786 
& 0.948 & 0.926 & 0.933
& 0.944 & 0.925 & 0.933
&\bf 0.959 &\bf 0.947 &\bf 0.952
& 0.943 & 0.912 & 0.921 \\
\bottomrule
\end{tabular}
}

\caption{Merged cross-validation results. P = Precision, R = Recall, F = F1-score. {\bf Bold} numbers are the best values and the {\underline{underlined}} numbers are the worst ones.}
\label{tab:merged_all}
\end{table*}

\begin{description}
    \item [TextCat.] The baseline obtains a better average value in Recall $\approx$86\% that in Precision$\approx$80\%. The variety {\bf HV} (Huasteca Veracruzana, best represented in \# of tokens), it is also the one most often confused with other varieties. {\bf HP} (Huasteca Potosina) --which is normal-- and {\bf CEA} (Mexicano del Centro Alto). The lowest-ranked class is {\bf HH} (Huasteca Hidalguense) with F1-Score$\approx$41\%.

\item [SVM.] The algorithm obtains good average values in Precision$\approx$95\% and Recall$\approx$93\%. The variety {\bf CEO} (Central de Occidente) is the best classified. The worst discriminated class is {\bf H} (Huasteca), with F1-Score$\approx$69\%.


\item[CNN.] The convolutional network obtains better average values in Precision$\approx$94\% than in Recall$\approx$93\%. The best identified variety is {\bf CEO} and {\bf H} the lowest ranked with F1-Score$\approx$63\%.


\item[LSTM.] The LSTM network achieves an average Precision$\approx$96\% and a Recall$\approx$95\%. The variety {\bf CEO} is the highest ranked and {\bf H} the lowest ranked with F1-Score$\approx$76\%. LSTM is the most efficient individual algorithm in the study.


\item[C-LSTM.] The Siamese network achieves an average Precision$\approx$95\% and an average Recall$\approx$94\%. The variety {\bf CEO} is the highest ranked and {\bf H} the lowest ranked with F1-Score$\approx$65\%.

\end{description}

\begin{figure}[h!]
\includegraphics[width=1.02\columnwidth]{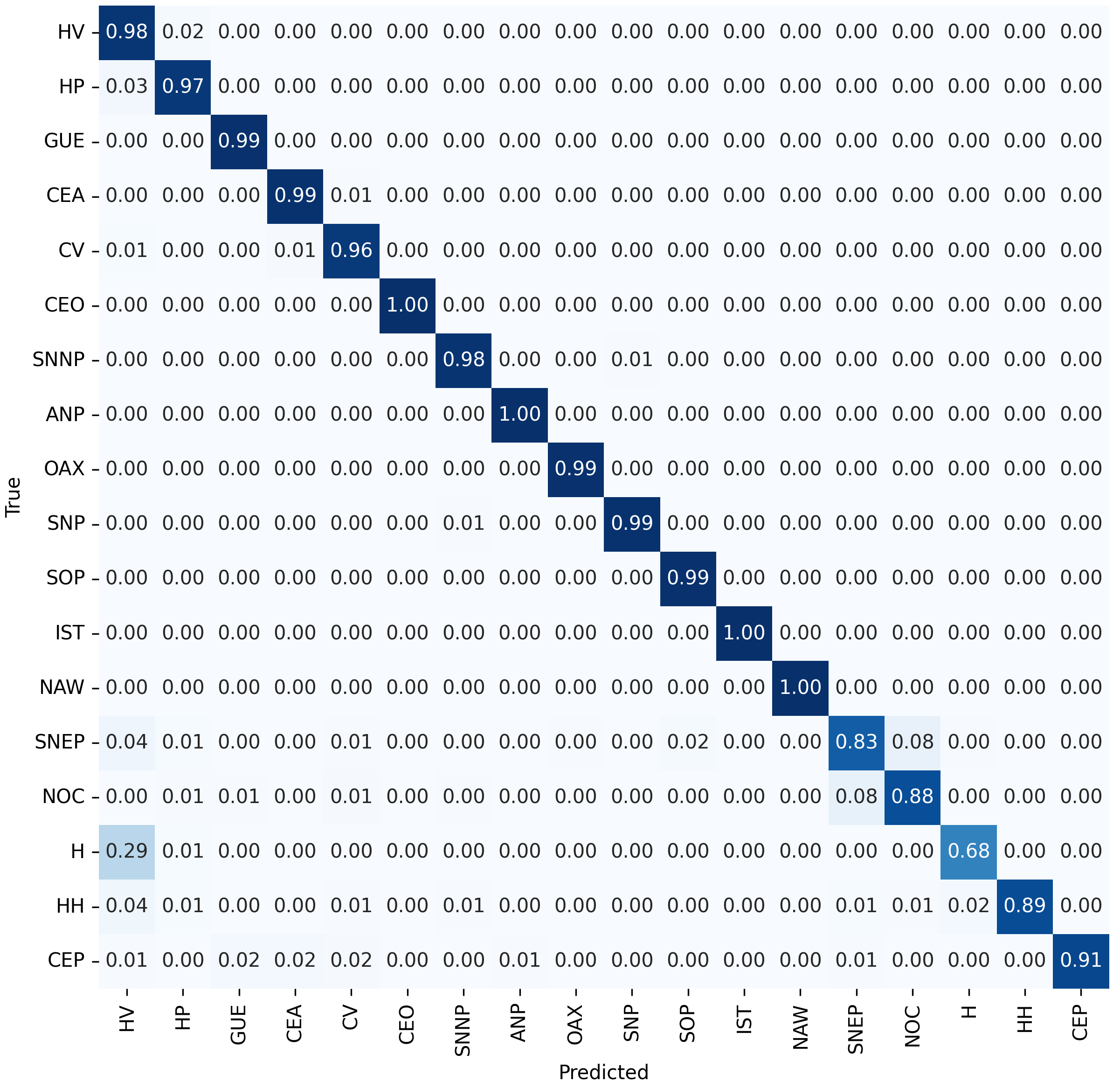}
 \caption{LSTM: Confusion matrix.}
 \label{fig:matrix_lstm_18}
\end{figure}

Finally, Figure \ref{fig:matrix_lstm_18} shows the confusion matrix for LSTM, the model having the better F1-score.

\section{Conclusions and Future Works}
\label{sec:conclusion}

The classification of dialectal varieties of Nawatl can be approached as a classic multi-class PLN problem.
It has been demonstrated that, although it is not a very simple task, our approach of combining representations and different learning algorithms achieved very good results.

Indeed, among the 18 varieties best represented in the corpus studied, the baseline of $2$ to $5$-grams character sets obtain acceptable results (F1-Score$\approx$79\%), far surpassed by neural methods: SVM and CNN (F$\approx$93\%), the C-LSTM Siamese network outperforms both (F1-Score$\approx$92\%), while the LSTM network achieved the best individual result (F1-Score$\approx$95\%). 

The confusion matrix shows excellent separation of the varieties studied. The most difficult variety to identify, the {\bf H} class --Huasteca Nawatl, that is a mix of varieties {\bf HV, HP} and {\bf HH}-- has now achieved F1-Score$\approx$74\%. 

The 95\% F1-Score obtained seems to be a very good result, but a 5\% error rate means 16 250 misclassified sentences, which is not insignificant even in this modestly sized corpus. Therefore, as part of our future work, we will conduct a voting process among the different models to see if the maximum value obtained can be improved with the LSTM model. This will allow us to determine whether using a method such as voting is superior to simply using the best model individually. We will also consider including other classification systems, probably based on linguistic rules, to help statistical or neural learning models better discriminate between classes.




\section*{Acknowledgments}
This work has been partially financed by Agorantic NAHU² projects and Intermedius PhD Grant, Avignon Université (France).

\bibliography{custom}

\begin{thebibliography}{36}
\providecommand{\natexlab}[1]{#1}

\bibitem[{Abdillahi et~al.(2006)Abdillahi, Nocera, and
  Torres}]{abdillahi:hal-01311495}
Nimaan Abdillahi, Pascal Nocera, and Juan~Manuel Torres. 2006.
\newblock \href {https://hal.science/hal-01311495} {{Boites a outils TAL pour
  les langues peu informatis{\'e}es : Le cas du Somali}}.
\newblock In \emph{{Journ{\'e}es d'Analyses des Donn{\'e}es Textuelles}},
  Besan{\c c}on, France.

\bibitem[{Albawi et~al.(2017)Albawi, Mohammed, and Al-Zawi}]{8308186}
Saad Albawi, Tareq~Abed Mohammed, and Saad Al-Zawi. 2017.
\newblock \href {https://doi.org/10.1109/ICEngTechnol.2017.8308186}
  {Understanding of a convolutional neural network}.
\newblock In \emph{International Conference on Engineering and Technology
  (ICET)}, pages 1--6.

\bibitem[{Berment(2004)}]{these-pi}
Vincent Berment. 2004.
\newblock \emph{Méthodes pour informatiser les langues et les groupes de
  langues ``peu dotées''}.
\newblock Ph.D. thesis, Université Joseph-Fourier - Grenoble I.

\bibitem[{Bochkarev et~al.(2012)Bochkarev, Shevlyakova, and
  Solovyev}]{DBLP:journals/corr/abs-1208-6109}
Vladimir~V. Bochkarev, Anna~V. Shevlyakova, and Valery~D. Solovyev. 2012.
\newblock \href {https://arxiv.org/abs/1208.6109} {Average word length dynamics
  as indicator of cultural changes in society}.
\newblock \emph{CoRR}, abs/1208.6109.

\bibitem[{Bojanowski et~al.(2017)Bojanowski, Grave, Joulin, and
  Mikolov}]{bojanowski-etal-2017-enriching}
Piotr Bojanowski, Edouard Grave, Armand Joulin, and Tomas Mikolov. 2017.
\newblock \href {https://aclanthology.org/Q17-1010} {Enriching word vectors
  with subword information}.
\newblock \emph{Transactions of the ACL}, 5:135--146.

\bibitem[{Canger(1988)}]{Canger1988NahuatlDA}
Una Canger. 1988.
\newblock \href {https://api.semanticscholar.org/CorpusID:144210796} {Nahuatl
  dialectology: A survey and some suggestions}.
\newblock \emph{International Journal of American Linguistics}, 54:28 -- 72.

\bibitem[{Cavnar et~al.(1994)Cavnar, Trenkle, and Mi}]{cavnar-1994-texcat}
W.B. Cavnar, J.M. Trenkle, and A.A. Mi. 1994.
\newblock \href
  {https://www.scirp.org/reference/referencespapers?referenceid=2641193}
  {N-gram-based text categorization}.
\newblock pages 161--175, Las Vegas, NV. Document Analysis and Information
  Retrieval.

\bibitem[{Cortes and Vapnik(1995)}]{SVM1995}
Corinna Cortes and Vladimir Vapnik. 1995.
\newblock Support-vector networks.
\newblock \emph{Machine learning}, 20(3):273--297.

\bibitem[{Francis-Landau et~al.(2016)Francis-Landau, Durrett, and
  Klein}]{francis-landau-etal-2016-capturing}
Matthew Francis-Landau, Greg Durrett, and Dan Klein. 2016.
\newblock \href {https://doi.org/10.18653/v1/N16-1150} {Capturing semantic
  similarity for entity linking with convolutional neural networks}.
\newblock In \emph{Conference of the North {A}merican Chapter of the
  Association for Computational Linguistics: Human Language Technologies},
  pages 1256--1261, San Diego, California. ACL.

\bibitem[{Gutierrez-Vasques et~al.(2016)Gutierrez-Vasques, Sierra, and
  Pompa}]{gutierrez2016axolotl}
Ximena Gutierrez-Vasques, Gerardo Sierra, and Isaac~Hernandez Pompa. 2016.
\newblock {Axolotl: a web accessible parallel corpus for Spanish-Nahuatl}.
\newblock In \emph{10th LREC'16}, pages 4210--4214.

\bibitem[{Guzm{\'a}n-Landa et~al.(2025{\natexlab{a}})Guzm{\'a}n-Landa,
  Torres-Moreno, Ranger, Garrido-Avenda{\~n}o, Figueroa-Saavedra,
  Quintana-Torres, Gonz{\'a}lez-Gallardo, Linhares-Pontes,
  Vel{\'a}zquez-Morales, and Moreno~Jim{\'e}nez}]{piyalliTALN}
Juan-Jos{\'e} Guzm{\'a}n-Landa, Juan-Manuel Torres-Moreno, Graham Ranger,
  Martha~Lorena Garrido-Avenda{\~n}o, Miguel Figueroa-Saavedra, Ligia
  Quintana-Torres, Carlos-Emiliano Gonz{\'a}lez-Gallardo, Elvys
  Linhares-Pontes, Patricia Vel{\'a}zquez-Morales, and Luis-Gil
  Moreno~Jim{\'e}nez. 2025{\natexlab{a}}.
\newblock \href {https://talnarchives.atala.org/TALN/TALN-2025/39.pdf}
  {$\pi$-yalli: un nouveau corpus pour le nahuatl / \textit{Yankuik
  nawatlahtolkorpus pampa tlahtolmachiotl}}.
\newblock In \emph{TALN Marseille}, pages 802--816. ATALA.

\bibitem[{Guzm{\'a}n-Landa et~al.(2025{\natexlab{b}})Guzm{\'a}n-Landa,
  Vázquez-Osorio, Torres-Moreno, Ranger, Garrido-Avenda{\~n}o,
  Figueroa-Saavedra, Quintana-Torres, Vel{\'a}zquez~Morales, and
  Sierra-Martínez}]{MICAI-piyalli-unigraph}
Juan-Jos{\'e} Guzm{\'a}n-Landa, Jesús Vázquez-Osorio, Juan-Manuel
  Torres-Moreno, Graham Ranger, Martha~Lorena Garrido-Avenda{\~n}o, Miguel
  Figueroa-Saavedra, Ligia Quintana-Torres, Patricia Vel{\'a}zquez~Morales, and
  Gerardo Sierra-Martínez. 2025{\natexlab{b}}.
\newblock A symbolic algorithm for the unification of nawatl word spellings.
\newblock In \emph{MICAI'25}, page 12p. SMIA.

\bibitem[{Hansen(2024)}]{hansen2024nahuatl}
Magnus~Pharao Hansen. 2024.
\newblock \emph{Nahuatl Nations: Language Revitalization and Semiotic
  Sovereignty in Indigenous Mexico}.
\newblock Oxford University Press.

\bibitem[{Hochreiter and Schmidhuber(1997)}]{LSTM}
Sepp Hochreiter and J\"{u}rgen Schmidhuber. 1997.
\newblock \href {https://doi.org/10.1162/neco.1997.9.8.1735} {Long short-term
  memory}.
\newblock \emph{Neural Compututation}, 9(8):1735–1780.

\bibitem[{Hornik et~al.(2013)Hornik, Mair, Rauch, Geiger, Buchta, and
  Feinerer}]{textcat}
Kurt Hornik, Patrick Mair, Johannes Rauch, Wilhelm Geiger, Christian Buchta,
  and Ingo Feinerer. 2013.
\newblock \href {https://doi.org/10.18637/jss.v052.i06} {The {textcat} package
  for $n$-gram based text categorization in {R}}.
\newblock \emph{Journal of Statistical Software}, 52(6):1--17.

\bibitem[{Huang et~al.(2022)Huang, Nie, and Huang}]{tsc-dpcnn}
Weichun Huang, Jiawei Nie, and Xiaohui Huang. 2022.
\newblock \href {https://doi.org/10.1051/itmconf/20224501040} {Text sentiment
  classification method based on dpcnn and bilstm}.
\newblock \emph{ITM Web of Conferences}, 45:01040.

\bibitem[{INEGI(2020)}]{inegi2020censo}
INEGI. 2020.
\newblock Censo de poblaci\'on y vivienda 2020.
\newblock
  \url{https://www.inegi.org.mx/rnm/index.php/catalog/632/study-description}.

\bibitem[{Jacovi et~al.(2018)Jacovi, Sar~Shalom, and
  Goldberg}]{jacovi-etal-2018-understanding}
Alon Jacovi, Oren Sar~Shalom, and Yoav Goldberg. 2018.
\newblock \href {https://doi.org/10.18653/v1/W18-5408} {Understanding
  convolutional neural networks for text classification}.
\newblock In \emph{{EMNLP} Workshop {B}lackbox{NLP}: Analyzing and Interpreting
  Neural Networks for {NLP}}, pages 56--65, Brussels, Belgium. ACL.

\bibitem[{Jamshidi et~al.(2024)Jamshidi, Mohammadi, Bagheri, Najafabadi,
  Rezvanian, Gheisari, Ghaderzadeh, Shahabi, and Wu}]{JAMSHIDI2024102306}
Saman Jamshidi, Mahin Mohammadi, Saeed Bagheri, Hamid~Esmaeili Najafabadi,
  Alireza Rezvanian, Mehdi Gheisari, Mustafa Ghaderzadeh, Amir~Shahab Shahabi,
  and Zongda Wu. 2024.
\newblock \href {https://doi.org/10.1016/j.datak.2024.102306} {Effective text
  classification using bert, mtm lstm, and dt}.
\newblock \emph{Data \& Knowledge Engineering}, 151:102306.

\bibitem[{Joachims(1998)}]{textCategorization_SVM}
Thorsten Joachims. 1998.
\newblock \href {https://doi.org/10.1007/BFb0026683} {Text categorization with
  support vector machines: learning with many relevant features}.
\newblock In \emph{10th European Conference on Machine Learning}, ECML'98, page
  137–142, Berlin, Heidelberg. Springer-Verlag.

\bibitem[{Johnson and Zhang(2017)}]{johnson-zhang-2017-deep}
Rie Johnson and Tong Zhang. 2017.
\newblock \href {https://doi.org/10.18653/v1/P17-1052} {Deep pyramid
  convolutional neural networks for text categorization}.
\newblock In \emph{55th Annual Meeting of the Association for Computational
  Linguistics (Vol 1: Long Papers)}, pages 562--570, Vancouver, Canada. ACL.

\bibitem[{Kim(2014)}]{kim-2014-convolutional}
Yoon Kim. 2014.
\newblock \href {https://doi.org/10.3115/v1/D14-1181} {Convolutional neural
  networks for sentence classification}.
\newblock In \emph{Conference on Empirical Methods in Natural Language
  Processing ({EMNLP})}, pages 1746--1751, Doha, Qatar. ACL.

\bibitem[{Lastra~de Su{\'a}rez(1986)}]{Lastra1986areas}
Yolanda Lastra~de Su{\'a}rez. 1986.
\newblock \emph{Las {\'a}reas dialectales del n{\'a}huatl moderno}.
\newblock UNAM, Instituto de Investigaciones Antropológicas, Mexico.

\bibitem[{Liang et~al.(2020)Liang, Zhu, Zhang, Cheng, and Jin}]{9407986}
Shengbin Liang, Bin Zhu, Yuying Zhang, Suying Cheng, and Jiangyong Jin. 2020.
\newblock \href {https://doi.org/10.1109/HPCC-SmartCity-DSS50907.2020.00169} {A
  double channel cnn-lstm model for text classification}.
\newblock In \emph{HPCC/SmartCity/DSS}, pages 1316--1321.

\bibitem[{Linhares~Pontes et~al.(2018)Linhares~Pontes, Huet, Linhares, and
  Torres-Moreno}]{siamese-TALN}
Elvys Linhares~Pontes, St{\'e}phane Huet, Andr{\'e}a~Carneiro Linhares, and
  Juan-Manuel Torres-Moreno. 2018.
\newblock \href {https://aclanthology.org/2018.jeptalnrecital-court.13}
  {Predicting the semantic textual similarity with {S}iamese {CNN} and {LSTM}}.
\newblock In \emph{TALN. Vol 1}, pages 311--320, Rennes, France. ATALA.

\bibitem[{Mairal et~al.(2014)Mairal, Koniusz, Harchaoui, and
  Schmid}]{10.5555/2969033.2969120}
Julien Mairal, Piotr Koniusz, Zaid Harchaoui, and Cordelia Schmid. 2014.
\newblock Convolutional kernel networks.
\newblock In \emph{28th International Conference on Neural Information
  Processing Systems - Vol 2}, NIPS'14, page 2627–2635, Cambridge, MA, USA.
  MIT Press.

\bibitem[{Manning and Schütze(1999)}]{Manning:99}
Christopher~D. Manning and Hinrich Schütze. 1999.
\newblock \emph{Foundations of Statistical Natural Language Processing}.
\newblock MIT Press, Cambridge, MA.

\bibitem[{Olko and Sullivan(2016)}]{olko2016bridging}
Justyna Olko and John Sullivan. 2016.
\newblock Bridging gaps and empowering speakers: An inclusive,
  partnership-based approach to nahuatl research and revitalization.
\newblock \emph{Integral strategies for language revitalization}, pages
  347--386.

\bibitem[{Paszke et~al.(2019)Paszke, Gross, Massa, Lerer, Bradbury, Chanan,
  Killeen, Lin, Gimelshein, Antiga, Desmaison, K\"{o}pf, Yang, DeVito, Raison,
  Tejani, Chilamkurthy, Steiner, Fang, Bai, and
  Chintala}]{10.5555/3454287.3455008}
Adam Paszke, Sam Gross, Francisco Massa, Adam Lerer, James Bradbury, Gregory
  Chanan, Trevor Killeen, Zeming Lin, Natalia Gimelshein, Luca Antiga, Alban
  Desmaison, Andreas K\"{o}pf, Edward Yang, Zach DeVito, Martin Raison, Alykhan
  Tejani, Sasank Chilamkurthy, Benoit Steiner, Lu~Fang, and 2 others. 2019.
\newblock \emph{PyTorch: an imperative style, high-performance deep learning
  library}.
\newblock Curran Associates Inc., Red Hook, NY, USA.

\bibitem[{Pugh and Tyers(2021)}]{Written-nahuatl-2021}
Robert Pugh and Francis Tyers. 2021.
\newblock \href {https://doi.org/10.18653/v1/2021.americasnlp-1.3}
  {Investigating variation in written forms of {N}ahuatl using character-based
  language models}.
\newblock In \emph{1st Workshop on Natural Language Processing for Indigenous
  Languages of the Americas}, pages 21--27, Online. ACL.

\bibitem[{Pugh and Tyers(2024)}]{Universal_Depend_pugh-tyers}
Robert Pugh and Francis Tyers. 2024.
\newblock \href {https://doi.org/10.18653/v1/2024.naacl-long.76} {A {U}niversal
  {D}ependencies treebank for {H}ighland {P}uebla {N}ahuatl}.
\newblock In \emph{Conference of the North American Chapter of the Association
  for Computational Linguistics: Human Language Technologies (Vol 1)}, pages
  1393--1403, Mexico. ACL.

\bibitem[{Zaman-Khan et~al.(2024)Zaman-Khan, Naeem, Guarasci, Khalid, Esposito,
  and Gargiulo}]{multiclass}
Haider Zaman-Khan, Muddasar Naeem, Raffaele Guarasci, Umamah~Bint Khalid,
  Massimo Esposito, and Francesco Gargiulo. 2024.
\newblock \href
  {http://dblp.uni-trier.de/db/journals/cys/cys28.html#ZamanKhanNGKEG24}
  {Enhancing text classification using bert: A transfer learning approach}.
\newblock \emph{Computación y Sistemas (CyS)}, 28(4).

\bibitem[{Zhang(2021)}]{9421225}
Yanbo Zhang. 2021.
\newblock \href {https://doi.org/10.1109/IPEC51340.2021.9421225} {Research on
  text classification method based on lstm neural network model}.
\newblock In \emph{2021 IEEE Asia-Pacific Conference on Image Processing,
  Electronics and Computers (IPEC)}, pages 1019--1022.

\bibitem[{Zhou et~al.(2015)Zhou, Sun, Liu, and Lau}]{CLSTM-Zhou}
Chunting Zhou, Chonglin Sun, Zhiyuan Liu, and Francis C.~M. Lau. 2015.
\newblock \href {https://arxiv.org/abs/1511.08630} {A {C-LSTM} neural network
  for text classification}.
\newblock \emph{CoRR}, abs/1511.08630.

\bibitem[{Zhou et~al.(2016)Zhou, Qi, Zheng, Xu, Bao, and Xu}]{tc-bi-lstm}
Peng Zhou, Zhenyu Qi, Suncong Zheng, Jiaming Xu, Hongyun Bao, and Bo~Xu. 2016.
\newblock \href {https://doi.org/10.48550/arXiv.1611.06639} {Text
  classification improved by integrating bidirectional {LSTM} with
  two-dimensional max pooling}.

\bibitem[{Zimmermann(2019)}]{zimmerman}
Klaus Zimmermann. 2019.
\newblock \href {https://doi.org/10.2436/rld.i71.2019.3255} {Estandarización y
  revitalización de lenguas amerindias: funciones comunicativas e
  ideológicas, expectativas ilusorias y condiciones de la aceptación}.
\newblock \emph{Revista de Llengua i Dret, Journal of Language and Law},
  71:111--122.

\end{thebibliography}


\end{document}